\documentclass[11pt,a4paper]{article}
\usepackage{times,latexsym}
\usepackage{url}
\usepackage[T1]{fontenc}

   \usepackage[acceptedWithA]{tacl2021v1}
\usepackage{tacl2021v1}

\usepackage{xspace,mfirstuc,tabulary}
\usepackage{graphicx}
\usepackage{boldline}
\usepackage{caption}
\usepackage{subcaption}
\usepackage{amsmath}
\usepackage{xparse}

\newif\iftaclinstructions
\taclinstructionsfalse 
\iftaclinstructions
\renewcommand{\confidential}{}
\renewcommand{\anonsubtext}{(No author info supplied here, for consistency with
TACL-submission anonymization requirements)}
\newcommand{\instr}
\fi

\iftaclpubformat 

\else

\fi

\NewDocumentCommand\emojibrain{}{{\includegraphics[width=3.8mm]{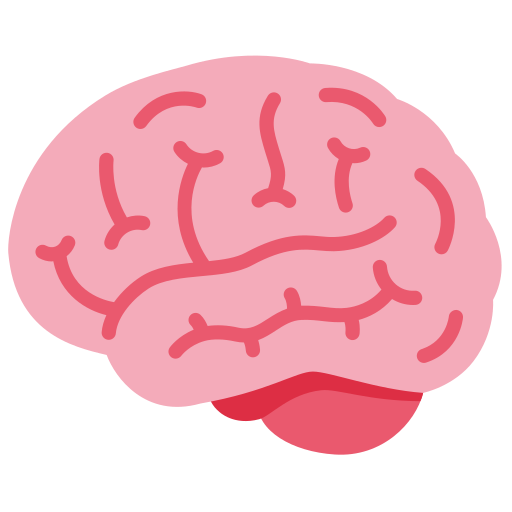}}{}}
\NewDocumentCommand\emojirobot{}{{\includegraphics[width=3.8mm]{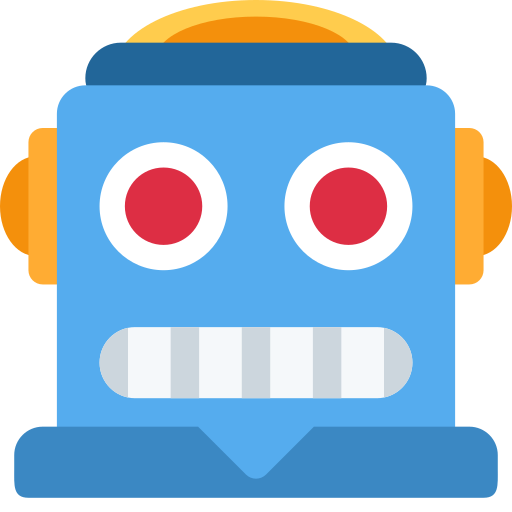}}{}}

\title{Automatic Scoring of Dream Reports’ Emotional Content with Large Language Models
}

\author{
  Lorenzo Bertolini$^{\emojirobot}$\quad Valentina Elce$^{\emojibrain}$ \quad Adriana Michalak$^{\emojibrain}$ \quad Giulio Bernardi$^{\emojibrain}$ \quad Julie Weeds$^{\emojirobot}$
  \\
  $^{\emojirobot}$Department of Informatics, University of Sussex, Brighton BN1 9RH, UK
  \\
  $^{\emojibrain}$MoMiLab Research Unit, IMT School for Advanced Studies, Lucca 55100, Italy
  \\
  \texttt{\{l.bertolini, juliewe\}@sussex.ac.uk}
  \\
  \texttt{\{valentina.elce, adriana.michalak, giulio.bernardi\}@imtlucca.it}
}

\date{}

\begin{document}
\maketitle
\begin{abstract}
In the field of dream research, the study of dream content typically relies on the analysis of verbal reports provided by dreamers upon awakening from their sleep. This task is classically performed through manual scoring provided by trained annotators, at a great time expense. While a consistent body of work suggests that natural language processing (NLP) tools can support the automatic analysis of dream reports, proposed methods lacked the ability to reason over a report's full context and required extensive data pre-processing. Furthermore, in most cases, these methods were not validated against standard manual scoring approaches. In this work, we address these limitations by adopting large language models (LLMs) to study and replicate the manual annotation of dream reports, using a mixture of off-the-shelf and bespoke approaches, with a focus on references to reports' emotions. Our results show that the off-the-shelf method achieves a low performance probably in light of inherent linguistic differences between reports collected in different (groups of) individuals. On the other hand, the proposed bespoke text classification method achieves a high performance, which is robust against potential biases. Overall, these observations indicate that our approach could find application in the analysis of large dream datasets and may favour reproducibility and comparability of results across studies.
\end{abstract}




\section{Introduction}
Dreams have fascinated humans since the dawn of time and their scientific study in the last decades even increased attention and interest towards this peculiar phenomenon. Indeed, available evidence suggests that dreams may be related to psychophysical well-being, and may be involved in or represent a window on sleep-dependent processes affecting the consolidation and integration of new memories \cite{Wamsley2011, Wamsley2014, dreaming_book}. Moreover, given their nature of internally generated conscious experiences, dreams (and their lack thereof) are regarded as a fundamental model to study and understand human consciousness \cite{Nir2010, Siclari2017}. In spite of this, the mechanisms that lead to dream generation and development, and the possible functions of dreams still remain poorly understood to this day. Among the factors that limit and slow down research on dreams is the fact that the content of dreams is difficult to assess quantitatively and in a reproducible way \cite{Elce2021}. 

\begin{figure}[htb!]
    \centering
    \includegraphics[width=\columnwidth]{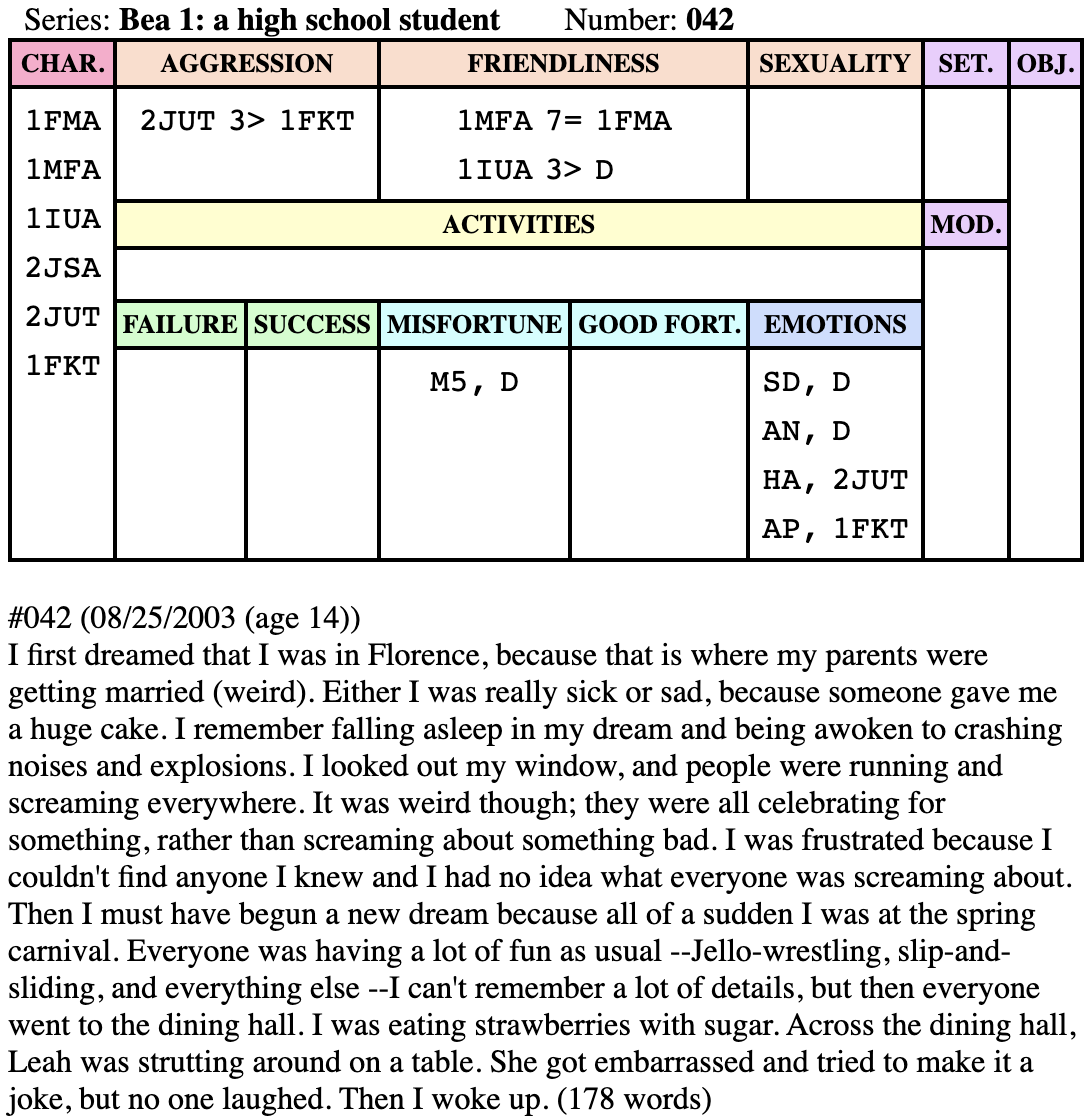}
    \caption{\label{fig:scoring_example}Example of a dream report with its annotation according to the Hall and Van de Castle (HVDC) scoring system.}
\end{figure}

Emotions are thought to represent an important component of most dream experiences. In fact, strong emotions are believed to facilitate dream recall, and experimental evidence suggests that dreams might have a direct role in regulating mood and emotional reactivity \citep{Blagrove2004}. Classically, the assessment of dream content --- including the presence of specific emotions --- is performed manually, by trained human operators, through the application of particular scales or scoring systems. While multiple scoring approaches exist to annotate and analyse dream reports, such as the scale by Hauri and colleagues \cite{hauri1975categorization}, or the rating system developed by Schredl \cite{Schredl_10}, the Hall and Van de Castle (HVDC) coding system \citep{Hall-1966-content} remains one of the most popular ones \cite{McNamara-etal-2019-Dream, Fogli2020}. An example of a dream annotated according to the HVDC system is presented in Figure \ref{fig:scoring_example}. When adopting this system, expert annotators first have to identify relevant characters (CHAR.) appearing in a dream, which might not necessarily include the dreamer themselves. Once identified, dreamer (D) and other characters are annotated with respect to relevant aspects (or features), such as activities, friendliness, and emotions.

A growing body of evidence has shown that NLP approaches can support the analysis of dream content through the automated quantification of references to specific words or topics (see \citet{Elce2021} for an extensive review). The adopted methods include in particular dictionary-based linguistic analysis \cite{Bulkeley2018, Mallett2021, Zheng2021} and distributional semantic models \cite{Razavi2013, Altszyler2017, Sanz2018}. Importantly, these approaches are based on the analysis of single words and thus have little-to-no access to the full content and context of each dream report, with a potential negative impact on performance. This may be especially relevant for references to emotions in dream reports, as the description of affective states may involve complex constructions and implicit information. In the present work, we propose to address these issues using pre-trained large language models (LLMs) based on Transformers \citep{vaswani2017attention}. Specifically, we investigate whether and how LLMs can support the detection and analysis of references to perceived emotions in dream reports, as defined in accordance with the HVDC framework. We first investigate if an off-the-shelf model could be used without further training or tuning to gather useful information on the ``general sentiment'' of a dream report, and how aligned its predictions are with respect to specific positive and negative emotions. We then study the ability of a fully-supervised solution, trained using previously scored dream reports, and propose additional experiments to test the trained model's robustness to different potential biases and out-of-distribution evaluation.

To the best of our knowledge, our work represents the first attempt to analyse and reproduce gold-standard HVDC annotations of dream reports with LLMs, and makes three main contributions. First, we show that, without direct supervision, the model’s predictions are largely unreliable, probably due to differences among reports collected from different individuals or groups of individuals. Second, we show how, despite the limited amount of training data, a fully-supervised approach based on multi-label text classification yields good and stable performance. Third, we provide follow-up experiments and analysis showing how the strategies learned by the model are robust with respect to out-of-distribution data, and those biases that affected the off-the-shelf approach.



\section{Related Work}

As summarised by \citet{Elce2021}, multiple pieces of evidence suggest that NLP methods can efficiently support the analysis of dream reports. However, of note, while emotions represent a fundamental component of oneiric experiences, only a few studies based on NLP methods explicitly and directly focused on the emotional aspects of dream reports \cite{nadeau-etal-2006-automatic, Razavi2013, McNamara-etal-2019-Dream, Yu2022}. From a general perspective, all of this work focused on two main solutions: dictionary-based linguistic analysis \cite{Bulkeley2018, Mallett2021, Zheng2021, Yu2022} and distributional semantic models \cite{nadeau-etal-2006-automatic, Razavi2013, Altszyler2017, Sanz2018, McNamara-etal-2019-Dream}. 

Dictionary-based methods analyse data word by word by comparing each item to a dictionary file that is structured as a collection of words defining different semantic categories. An example could be the `positive emotion' category, containing a series of words such as \textit{``joy’’}, \textit{``happiness’’}, and \textit{``smiling’’}. Approaches based on these methods \cite{Bulkeley2018, Mallett2021, Zheng2021, Yu2022} are mainly used to determine the relative frequency of references to specific contents within textual data, and can hence be inherently misleading since they do not account for the context or the syntactic structure. For instance, consider a sentence like \textit{``I should have felt sadness but I did not.''}. A dictionary-based approach may recognize the word \textit{``sadness’’} as indicating a negative emotion thus providing a misleading interpretation of the actual described affective state. 

Computational approaches that used distributional semantics \cite{nadeau-etal-2006-automatic, Razavi2013, Altszyler2017, Sanz2018, McNamara-etal-2019-Dream} were largely based on word-level representation obtained using models like word2vec \cite{mikolov-etal-2013-efficent}, latent semantic analysis (LSA) \cite{Landauer1997AST}, or GloVe \cite{pennington-etal-2014-glove}. In such cases, the encodings of textual reports were generated by adding or averaging word-level embeddings. These methods may perform better than dictionary-based methods at capturing the semantic content of a report as they include partial information about the ‘context’ in which words are commonly used and can rely on continuous relative distances between words rather than on dichotomous decisions. However, these approaches had no access to syntactic structure and more contextual understanding, since sentence and document-level encodings were based on vector average or addition \cite{klafka-ettinger-2020-spying}. In other words, these methods would generate identical latent representations for sentences like \textit{``The angry monster was running towards the happy boy.''} and \textit{``The angry boy was running towards the happy monster.''}.

Although with similar limitations, a niche of previous work applied a combination of NLP and machine learning methods to assess specific emotional aspects of dream reports \cite{Razavi2013, McNamara-etal-2019-Dream, Yu2022}. \citet{Razavi2013} combined ad-hoc classifiers with a distributional-based approach to detect potential shifts in sentiment within each report. Evaluated reports were extracted from a large public database (DreamBank), and were (re-)scored by the authors using a four-level emotional rating (very-negative to very-positive). Despite relying on a mixture of local (word-to-word) and general (sentence-to-sentence) occurrences, the adopted approach strongly relies on extensive data pre-processing, as well as composition by averaging, hence losing access to structural and deeper semantic information. \citet{McNamara-etal-2019-Dream} used a pre-trained agent to detect recurrent themes in a series of reports and found a partial match in the retrieved themes with aspects of the HVDC system as well as significant differences in how themes occurred in male vs. female dreamers. The distribution of these themes was then used to assess the `mood' of each report. \citet{Yu2022} combined a dictionary-based method with support vector machines \cite{cortes1995support} (SVM) to describe the general sentiment of dream reports in multiple languages. 
Overall, the described studies present two main differences with respect to our work. First, the models employed in those studies lacked access to the global context of each report. Second, the annotations and evaluations of emotional states were not directly compared against widely used coding systems such as the HVDC. In our work, we propose to use a pre-trained large language model (LLM) to encode dream reports and thus allow the model to have full access to the context. Furthermore, our analyses compared the results with gold-standard annotations from the HVDC system, thus highlighting the possible value of our approach as a replacement or support for manual annotations in dream research.




\section{Dataset}\label{sec:task_benchmark}
For our experiments, we utilise a subset of reports extracted from the  DreamBank database\footnote{\url{https://www.dreambank.net}} \cite{Domhoff2008}, pre-annotated according to the Hall and Van De Castle (HVDC) coding system \citep{Hall-1966-content}. \texttt{DreamBank.net} consists of a collection of over 20K dream reports gathered from different sources and organised in series, either provided by single individuals or groups of people, such as college students, teenagers and blind adults. While \texttt{DreamBank.net} can be freely explored, the reports and the HVDC scores adopted in the current work are made available upon direct request to the researchers who maintain the DreamBank website. Among the approximately 1.8K labelled dream reports, all in the English language, 922 contained at least one emotion associated with either the dreamer or another character. Given the scope of this work, we focused our experiments on those reports containing at least one emotion (n=922). The dataset can be further divided into six series: \texttt{Bea 1: a high school student} (n=171/99; total number of reports/reports including at least one emotion), \texttt{Ed: dreams of his late wife} (n=143/108), \texttt{Emma: 48 years of dreams} (n=300/81), \texttt{Hall/VdC Norms: Female} (n=491/280), \texttt{Hall/VdC Norms: Male} (n=500/203), \texttt{Barb Sanders: baseline} (n=250/151). 

\begin{figure}[htb!]
    \centering
    \includegraphics[width=\linewidth]{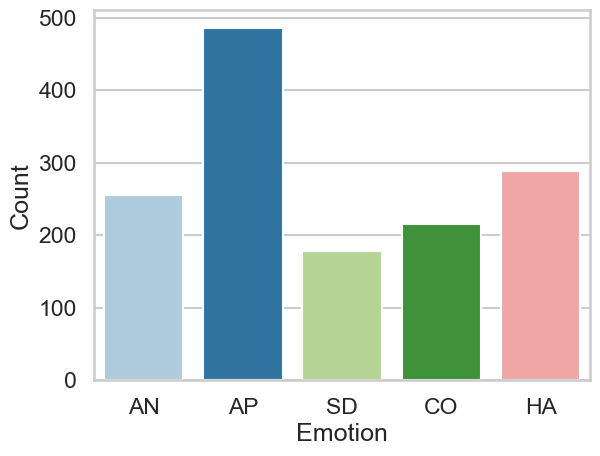} \caption{\label{fig:db_em_dist} General emotion distribution across DreamBank.}
\end{figure}

The HVDC coding system examines ten categories of elements appearing in dream reports (characters, interactions, emotions, activities, striving, (mis)fortunes, settings and objects, descriptive elements, food and eating, elements from the past). Within this study, we focused only on the annotation of the emotions. In the HVDC coding system, emotions are divided into 5 classes, that are anger (AN), sadness (SD), apprehension (AP), confusion (CO), and happiness (HA). Emotions might be assigned either to the dreamer or to other dream characters. Here, we analysed both the emotions scored as experienced by the dreamer (\textit{Dreamer Emotions}) and the occurrence of references to emotions in the dreams, regardless of the dream characters who experienced them (\textit{General Emotions}). An example of such distinction is given in Figure \ref{fig:usa_GS_exp}. Of note, according to the HVDC system, multiple references to the same emotion within the same report are treated separately. Figure \ref{fig:db_em_dist} summarises how single emotions are distributed in the dataset. Of note, under both Dreamer and General Emotions, the majority of reports contain only one emotion (circa 75\% and 65\%, respectively), while almost the whole dataset can be represented by reports containing up to two emotions (approximately 95\% and 90\%, respectively). See Appendix \ref{app_DB_dist} for more details). All the code used to run the experiments and analyses are available here\footnote{\url{https://github.com/lorenzoscottb/Dream_Reports_Annotation}}.

%



\section{Off-the-Shelf Sentiment Analysis}\label{sec:usa}
\begin{figure*}[htb!]
    \includegraphics[width=\textwidth]{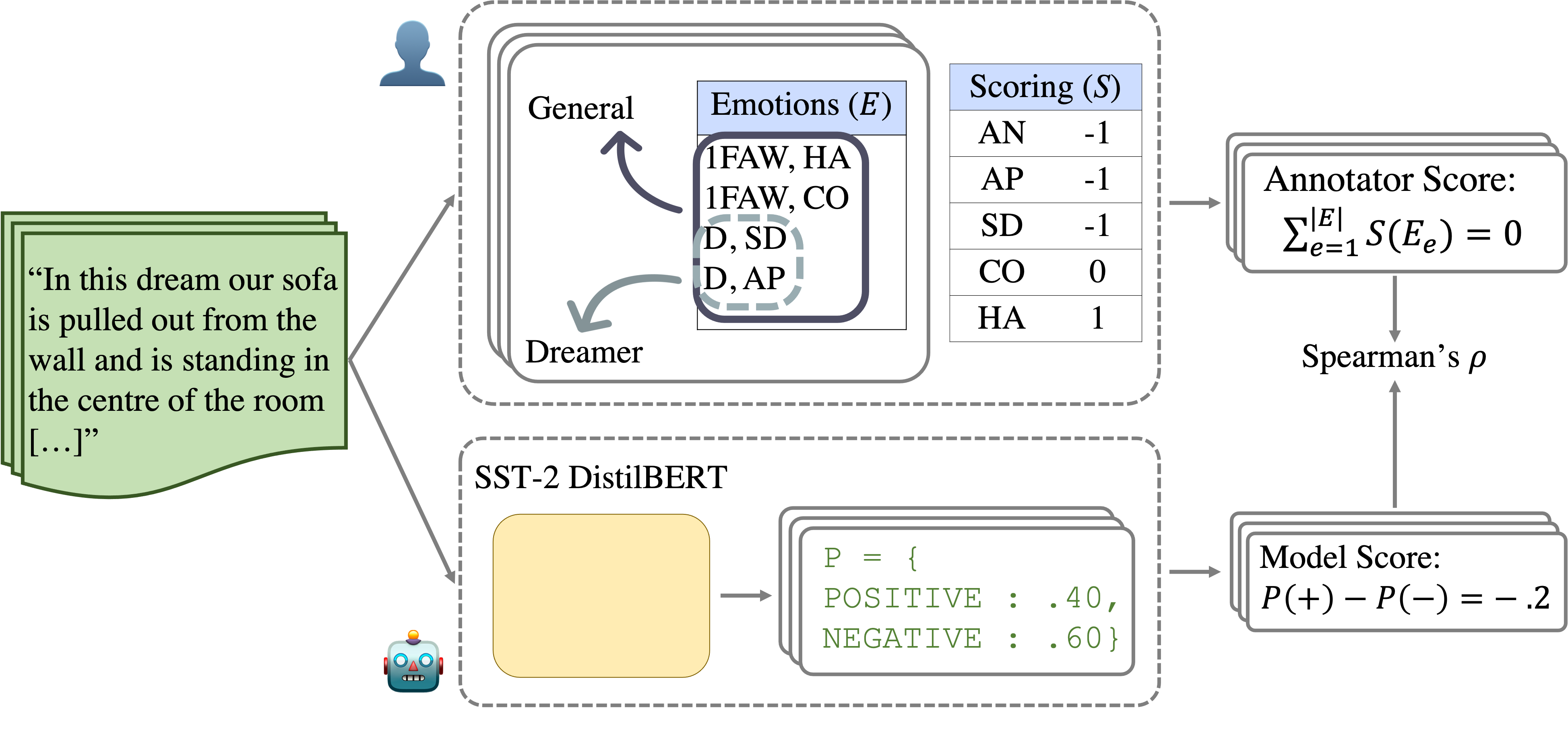}
    \caption{\label{fig:usa_GS_exp} Proposed setup for the Report Sentiment experiment (Section \ref{sec:usa_RS}).}
\end{figure*}

First, we investigate if an off-the-shelf model tuned to perform sentiment analysis could be used to assess the emotional content of dream reports. Specifically, we propose to test a 2-way \texttt{POSITIVE} vs. \texttt{NEGATIVE} classification, similar to previous work \cite{McNamara-etal-2019-Dream, Yu2022}. To ensure both usability and replicability, the experiment was run via Hugging Face's \cite{wolf-etal-2020-transformers} \texttt{pipeline} functionality, using the default setting for the \texttt{sentiment-analysis} task\footnote{That is, a DistilBERT model tuned on SST-2 \cite{socher-etal-2013-recursive}}. Specifically, we adopt a two-level experiment. In the first experiment, we investigate whether the general predictions of the model (i.e., the predicted labels and their scores) correlate with the \textit{sentiment} of individual dream reports. To this aim, we defined the overall sentiment of each report as the sum of all references to emotions identified according to the HVDC coding system. A schematic summary of our approach is presented in Figure \ref{fig:usa_GS_exp}.

The second experiment focuses on those reports containing a single emotion, and studies whether the predicted label (i.e, \texttt{POSITIVE} or \texttt{NEGATIVE}) matches the emotion found by the annotators.

\subsection{Annotator Score}\label{sec:usa_RS}
While 90\% of the gathered Dream reports don't contain more than two emotions some reports can contain a large variety of emotions --- up to 9 in some rare cases (see Appendix x \ref{app_DB_dist} for more details). hence, the main aim of the sentiment analysis investigation is to assess whether the model's predictions do reflect the overall \textit{sentiment} of a report, defined according to the number of times specific references to positive or negative emotions appear in a report (regardless of the character who experienced them). More formally, given a dream report containing a list of Emotions $E$, such as the one in the example of Figure \ref{fig:usa_GS_exp}, and scoring table $S$, mapping each HVDC emotion to an integer value, we compute the sentiment of a report (i.e., the \textit{Annotator Score}) through the equation in \ref{eq:gen_sent}

\begin{equation}
\sum_{e=1}^{|E|}S(E_{e})
\label{eq:gen_sent}
\end{equation}

For this experiment, our scoring table $S$ assigns to \textit{anger}, \textit{apprehension}, \textit{sadness} the value of -1, \textit{happiness} has a positive value of 1, and \textit{confusion} is treated as neutral, and hence has the value of 0 (see Figure \ref{fig:usa_GS_exp}). The \textit{Model Score} of a report is defined as the difference between the probability associated with the \texttt{POSITIVE} and \texttt{NEGATIVE} labels (see Figure \ref{fig:usa_GS_exp}). For instance, if the model predicted the probability distribution of the \texttt{POSITIVE} ($P(+)$) and \texttt{NEGATIVE} ($P(-)$) labels to be .4 and .6, respectively, then the \textit{Model Score} for such a report would be –.2. 

For the current experiment, the model’s performance is assessed by comparing the \textit{Model Score} with the \textit{Annotator Score} via Spearman's correlation coefficient ($\rho$). For both the current and following experiments, we will compute two distinct sets of model and human-generated values, based on whether the evaluation process will make use solely of those emotions referring to the dreamer (or Dreamer Emotions), or the full list of emotions (i.e., the General Emotions), regardless of whether they were associated with the dreamer or other characters. An example of such distinction can be found in Figure \ref{fig:usa_GS_exp}.

\subsubsection{Results} 
\begin{figure}[htb!]
    \centering
    \includegraphics[width=\columnwidth]{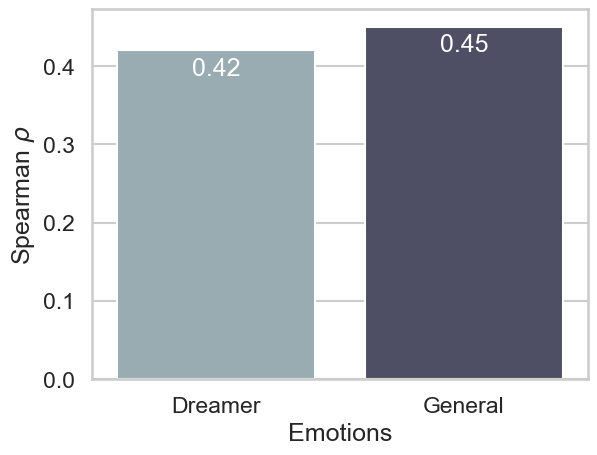}
    \caption{\label{fig:general_sentiment_main} Annotator score results. Correlations coefficients (in Spearman's $\rho$) between the model-predicted scores and the reports' annotator scores.}
\end{figure}

\begin{figure}[htb!]
    \centering
    \includegraphics[width=\columnwidth]{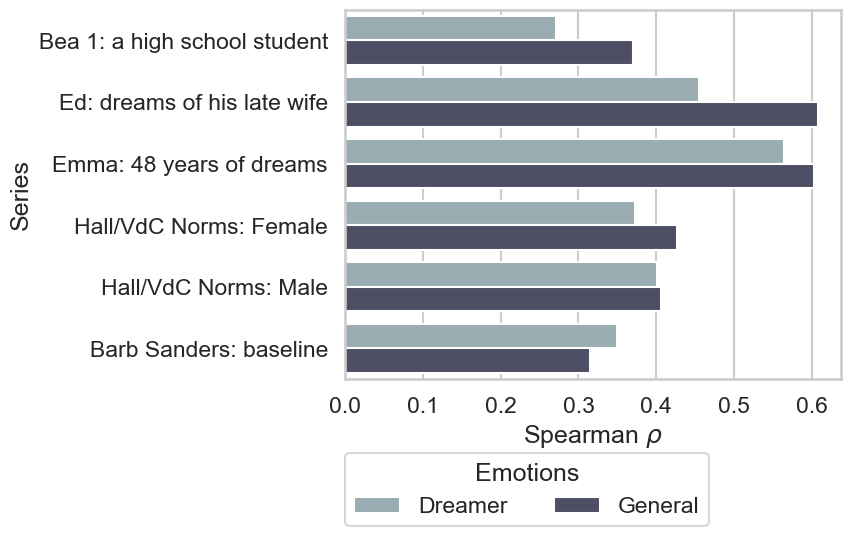}
    \caption{\label{fig:general_sentiment_correlations}Annotator score, collection analysis. Correlations coefficients (in Spearman's $\rho$) between the model-predicted and annotator-produced reports' scores and each r, divided by DreamBank's Series.}
\end{figure}

Figure \ref{fig:general_sentiment_main} presents the results of the correlation analysis between scores produced by human annotators and the selected model, divided by Dreamer and  General Emotions. While the correlation with the General Emotions is marginally better, results are overall poor. Moreover, under both Dreamer and General Emotions, the performance is heavily influenced by different DreamBanks' series, as demonstrated by Figure \ref{fig:general_sentiment_correlations}. Interestingly, under both the Dreamer and the General Emotions, \texttt{Ed} and \texttt{Emma} show the strongest correlation between human and model scores.

\begin{figure}[htb!]
    \centering
    \includegraphics[width=\columnwidth]{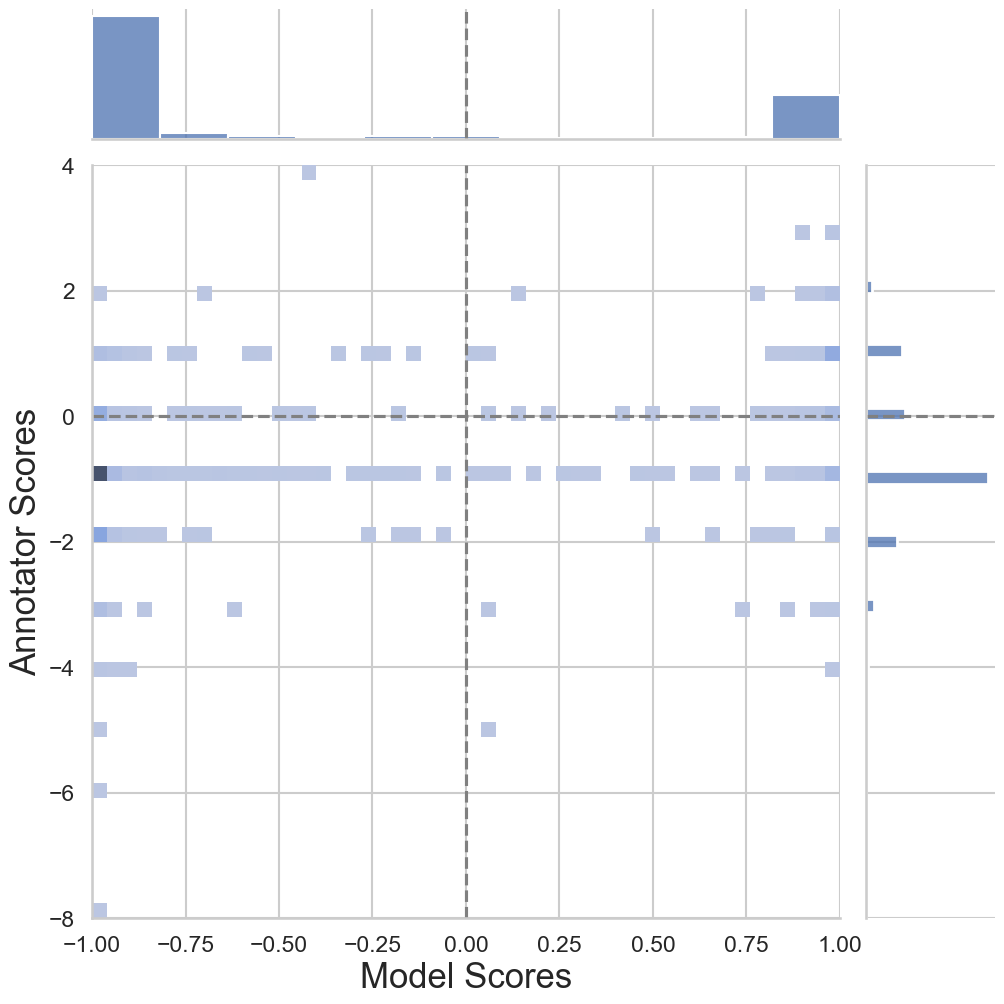}
    \caption{\label{fig:general_sentiment_dist} Annotator Score, predictions' analysis. Comparison of the Predicted Sentiment scores (x axis) and Report Sentiment distribution (y axis) for the General Emotion set. As seen, while the model's predicted scores are strongly polarised, annotators' scores, computed via Eq. \ref{eq:gen_sent}, are more smoothly and evenly distributed.}
\end{figure}

Figure \ref{fig:general_sentiment_dist} suggests that these results are likely due to the remarkably different distributions produced by human annotators and the sentiment analysis model. While the (General) \textit{Annotator Scores} (y axis) are unimodally (and almost normally) distributed, the predictions of the model (i.e., the \textit{Model Scores}, x axis) are strongly polarised. In other words, the model was consistently very confident in its decisions on which sentiment (\texttt{POSITIVE} or \texttt{NEGATIVE})) was appearing in a given report. Interestingly, the peak of the \textit{Annotator Scores} is around the score of –1, which coincides with a portion of the model's predictions that spans throughout the whole interval (although with a slightly higher concentration around – 1). As presented in more detail in Appendix \ref{app_hists}, only a very small percentage of this variance of \textit{Model Scores} could be potentially attributed to the model focusing on one of multiple emotions and ignoring the resent, since the vast majority of these reports with an \textit{Annotator Score} of –1 only contain one emotion\footnote{Of note, a similar argument can be made for reports containing two emotions. The vast majority of those reports are in fact clustered around \texttt{Annotator Scores} of –2 and 2, hence containing only negatively or positively connoted emotions (see Appendix \ref{app_hists} for more details)}. This might suggest that the score of the model can not efficiently reflect a more general sentiment, but only encode the presence of a specific type (positive or negative) emotions. The following experiment will investigate this possibility, focusing on those reports only containing one emotion, and approaching the problem from a categorical perspective.

\subsection{Single-Emotion}\label{sec:usa_EA}
The first experiment showed how the selected model fails to correctly capture the distribution of human annotators' scores, mainly due to very polarised predictions, and might simply reflect what type of emotion (positive or negative) is mainly present in a given report. Since the HVDC system also allows to assign a strictly positive or negative connotation to each emotion, we study such possibility by focusing solely on those reports that experts have annotated with one --- and only one --- of the five HVDC emotions: \textit{anger}, \textit{apprehension}, \textit{confusion}, \textit{sadness} and \textit{happiness}. The goal is thus to understand if reports classified as \texttt{POSITIVE} or \texttt{NEGATIVE} by the model do contain an emotion that the HVDC scoring system also defined as positive or negative. Here, results will be interpreted in terms of precision, recall and F1, with respect to the two prediction classes (\texttt{POSITIVE} and \texttt{NEGATIVE}).

\begin{table*}[htb!]
\centering
\small
\resizebox{\textwidth}{!}{%
\begin{tabular}{lcccccc}
\hlineB{3}
                     & \multicolumn{2}{c}{\textbf{Precision}} & \multicolumn{2}{c}{\textbf{Recall}} & \multicolumn{2}{c}{\textbf{F1}} \\
\multicolumn{1}{l}{} & Dreamer &     General &      Dreamer &      General &      Dreamer &     General \\ 
\hlineB{3}
\texttt{NEGATIVE}     & 92 & 91   &  83 & 82  &  87 & 86 \\
\texttt{POSITIVE}     & 44 & 45   &  64 & 65  &  52 & 53 \\ \hline
macro avg             & 68 & 68   &  73 & 73  &  70 & 70 \\
weighted avg          & 83 & 82   &  79 & 78  &  81 & 80 \\
\hlineB{3}
\end{tabular}
}
\caption{\label{tab:Single_emotion_f1}Single-emotion results. Per-class and average scores obtained when comparing model-predicted and human-generated labels for dream reports containing a single emotion. Here, the five HVDC emotions were collapsed into positive (i.e., \textit{happiness}) and negative (i.e., \textit{anger}, \textit{apprehension}, \textit{sadness} and \textit{confusion}), and compared against the label predicted by the sentiment analysis model (i.e., \texttt{POSITIVE} or \texttt{NEGATIVE}).}
\end{table*}

\begin{figure*}[htb!]
    \centering
    \includegraphics[width=\textwidth]{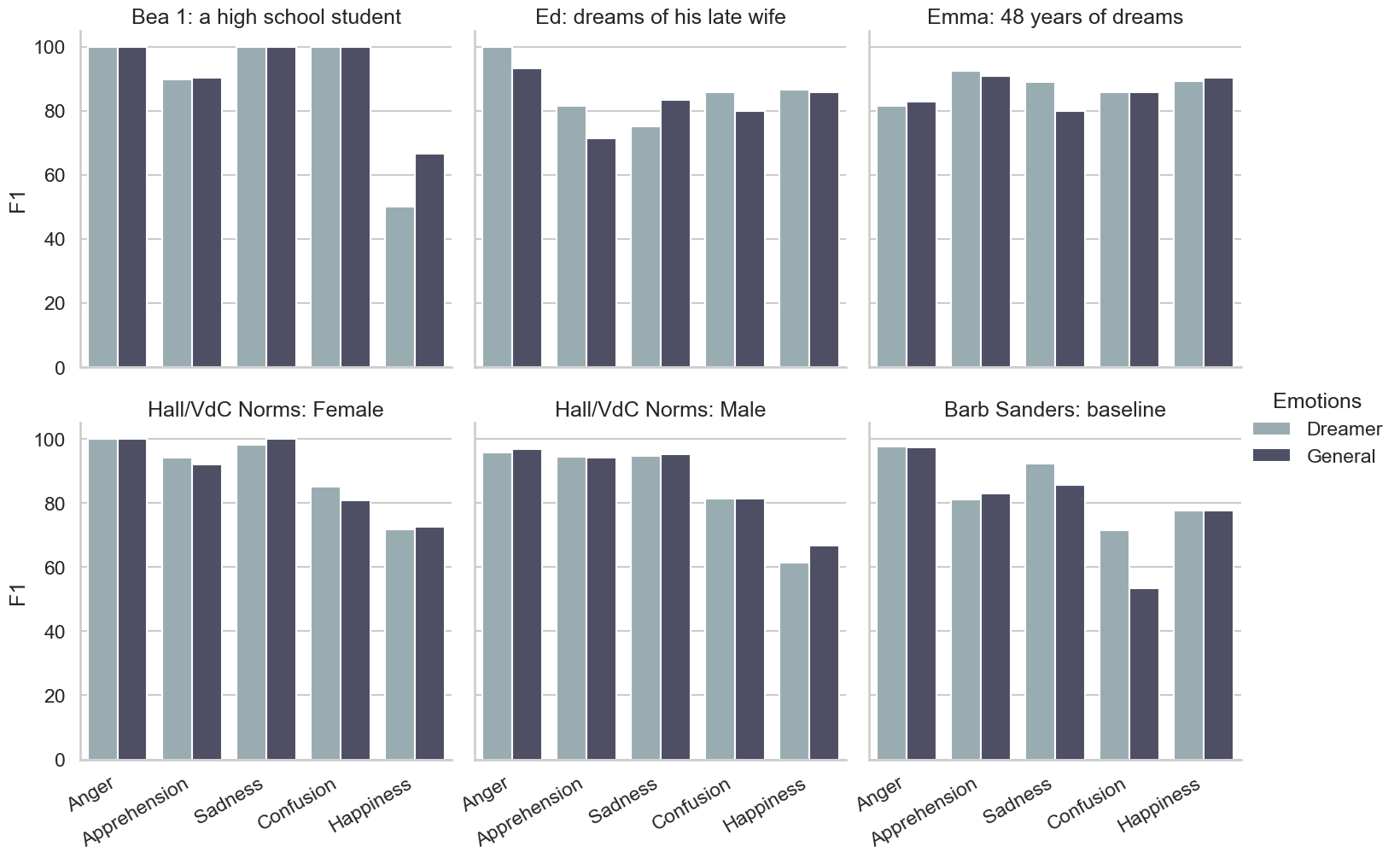}
    \caption{\label{fig:usa_collection} Single-emotion: Series and emotions analysis. Results (in terms of reference-class F1 scores) obtained by the model for each HVDC emotion (x axis), DreamBank's series  (diagrams), and Emotions (Dreamer vs. General, hue). For \textit{happiness}, the F1 scores reference class is \texttt{POSITVE}, while all other HVDC emotions share \texttt{NEGATIVE} as their reference-class for the reported F1 scores.}
\end{figure*} 

\subsubsection{Results} 
Table \ref{tab:Single_emotion_f1} summarises the results and clearly shows that, with respect to reports containing a single emotion, the predictions of the model match the human-produced annotations only with respect to negative emotions, while showing poor results with respect to the \texttt{POSITIVE} class ––– which only contains \textit{happiness}. The model is however largely unstable with respect to the type of error it makes, as shown by the notable difference between precision and recall scores. 

Figure \ref{fig:usa_collection} presents the same results of Table \ref{tab:Single_emotion_f1}, divided by single HVDC emotion (x axis) and series (diagrams), and shows how the model remains notably impacted by the different DreamBank's series. Of note, \texttt{Ed} and \texttt{Emma}, the two series that produced the best performance in the previous experiment, here show the most balanced results across different HVDC emotions. Overall, these results strongly suggest that the selected model had fewer problems when classifying reports containing negative emotions than at detecting the presence of positive emotions. 




\begin{figure*}[tbhp]
\centering
\includegraphics[width=\linewidth]{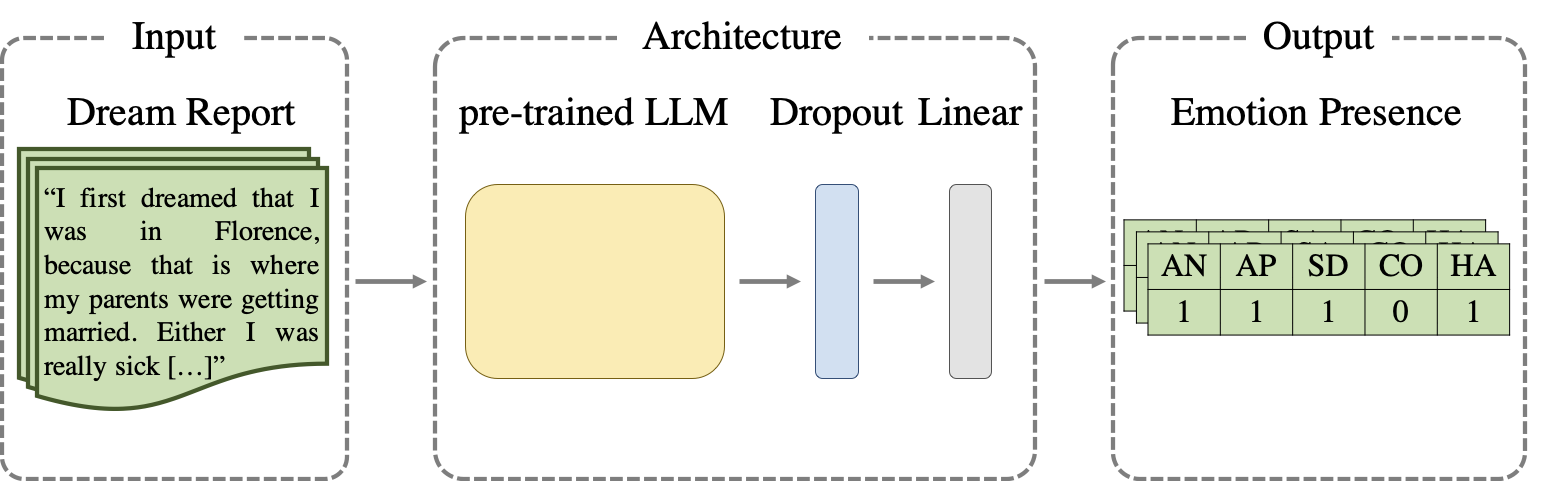}
\caption{Schematic view of the adopted architecture and training procedure for the bespoke multi-label classification experiments. Given a set of dream reports from DreamBank as input, the architecture is trained end-to-end to guess which of the five emotions recognised by the Hall and Van de Castle (HVDC) system --- \textbf{anger (AN)}, \textbf{apprehension (AP)}, \textbf{sadness (SD)}, \textbf{confusion (CO)}, and \textbf{happiness (HA)} --- is present (1) or absent (0) in each report. The adopted architecture is constructed out of three components: a pre-trained LLM (in our case, a BERT-large-cased model), a dropout layer, and a linear layer.}
\label{fig:classification}
\end{figure*}

\section{Bespoke Multi-Label Classification}\label{sec:classification}
The previous section showed that the selected off-the-shelf sentiment analysis model failed to coherently reproduce the distribution of the expert-annotator scores and that the model prediction where more reliable with respect to negative emotions. This appears to be at least in part due to relevant differences in performance across different subsets (or series) of DreamBank. Therefore, we next investigate whether these problems could be addressed using a bespoke text classification model, trained directly on the gold-standard HVDC annotations. In this case, rather than limiting our task to predicting the sentiment polarity of each report (i.e., \texttt{NEGATIVE} vs \texttt{POSITIVE}), we attempted a more fine-grained classification, aimed at determining the presence (1) or absence (0) of distinct emotions (\textit{anger} (AN), \textit{sadness} (SD), \textit{apprehension} (AP), \textit{confusion} (CO), \textit{happiness} (HA)), regardless of the number of times they appear in a given report. The task is hence formulated as a multi-label classification, in which a model is trained to simultaneously and independently predict \textit{if} each of the emotions was identified by expert annotators in each dream report.

\subsection{Experimental Setup}
Similarly to Section \ref{sec:usa}, the current experiments are primarily based on a BERT encoder, implemented via Hugging Face. However, the LLM is in this case integrated into a three-component architecture, summarised in Figure \ref{fig:classification}. The first component is a BERT-large-cased model, used to obtain the encoding of each report, by extracting the final layer's \texttt{[CLS]} vector. Encodings are then fed to a dropout layer (with $p=.3$) and a linear layer, reducing the number of the dimensions to the number of desired classes, corresponding to the five HVDC emotions. The described architecture is then trained end-to-end with a binary cross-entropy loss, with the addition of a sigmoid layer between the loss and the linear layer, and adopting a $K$-fold cross-validation procedure (with $K$=5). At each fold, the dataset is divided into 80\% for training and 20\% for testing, and the architecture is trained for 10 epochs, using dream reports are used as input, and the presence of HVDC emotions as the output to predict. While the previous section was evaluating a model on the HVDC annotation indirectly (i.e., by assigning each emotion in a report a score, and summing those scores, see Figure \ref{fig:usa_GS_exp}) we here \textit{directly} evaluate the model in the HVDC's gold-starred annotation framework, by training and testing the model to simultaneously and independently guess if each of the five HVDC emotions was annotated as present by the expert annotators (see Figure \ref{fig:classification}. Hence, similarly to previous work investigating the presence of emotions in dream reports, we adopted precision, recall, and F1 as evaluation metrics \cite{Fogli2020}.

\begin{table*}[htb!]
\centering
\resizebox{\textwidth}{!}{%
\begin{tabular}{lcccccc}
\hlineB{3}
                     & \multicolumn{2}{c}{\textbf{Precision}} & \multicolumn{2}{c}{\textbf{Recall}} & \multicolumn{2}{c}{\textbf{F1}} \\
\multicolumn{1}{l}{} & Dreamer &     General &      Dreamer &      General &      Dreamer &     General \\ 
\hlineB{3}
Anger (AN)           &   86 ± 9 &  85 ± 7 &   89 ± 3 &   89 ± 4 &   87 ± 4 &  87 ± 5 \\
Apprehension (AP)           &   86 ± 4 &  88 ± 7 &   88 ± 5 &   92 ± 3 &   87 ± 3 &  89 ± 3 \\
Sadness (SD)           &  84 ± 10 &  84 ± 4 &  72 ± 15 &  77 ± 12 &  77 ± 11 &  80 ± 7 \\ 
Confusion (CO)           &   90 ± 5 &  92 ± 2 &   76 ± 6 &   85 ± 5 &   82 ± 5 &  88 ± 3 \\
Happiness (HA)           &   93 ± 5 &  86 ± 4 &   85 ± 6 &   88 ± 3 &   89 ± 5 &  87 ± 2 \\ \hline
macro avg    &   88 ± 3 &  87 ± 3 &   82 ± 5 &   86 ± 2 &   85 ± 3 &  86 ± 2 \\
micro avg    &   87 ± 3 &  87 ± 3 &   84 ± 4 &   87 ± 2 &   85 ± 3 &  87 ± 2 \\
samples avg  &   88 ± 2 &  89 ± 2 &   87 ± 3 &   90 ± 2 &   86 ± 3 &  88 ± 2 \\
weighted avg &   88 ± 2 &  87 ± 4 &   84 ± 4 &   87 ± 2 &   85 ± 3 &  87 ± 2 \\
\hlineB{3}
\end{tabular}
}
\caption{\label{tab:multi_class_kfold_overall} Bespoke muli-label classification results. Average scores (± standard deviation) of the 5-fold cross-validation text classification experiment.Dreamer and General columns refer to the Emotions used for training and testing. While under the General Emotions setting we made use of all emotions found by expert annotators in each report (see Figure \ref{fig:scoring_example}), the Dreamer Emotions refers to the subset of the General Emotions associated by the expert annotators solely to the dreamer (D in Figure \ref{fig:scoring_example}).}
\end{table*}

\subsection{Results}
Table \ref{tab:multi_class_kfold_overall} summarises the average scores (± standard  deviations) obtained by the architecture for Dreamer and General emotions. Overall, F1 scores show a strong and generally stable performance. The minimal difference between macro and weighted F1 scores further suggests that the difference in support instance only has a marginal impact. With respect to single Emotion sets, performance tended to be higher and more stable for General than for Dreamer emotions. When trained and tested on General emotions, the models show a notable balance between precision and recall, despite a relatively higher variance across precision measures. On the other hand, models trained solely with Dreamer emotions present an overall higher precision than recall, with the latter being notably less stable. These patterns are likely explained by the low number of emotions-per-report associated with the Dreamer set, while the emotion distribution is more balanced in the General set. Models trained solely with the Dreamer set are hence notably less prone to produce False-Positive errors but produce a higher amount of False-Negative errors\footnote{Remember that precision is computed as TP/(TP+FP), while recall is equal to TP/(TP+FN)}. Since the General emotion set is overall more  balanced, the models' performance is higher and more stable across precision and recall. 

With respect to single emotions, it is difficult to identify a shared pattern of results. A notable exception is \textit{sadness} (SD). Under both emotion sets, models appear to notably struggle to classify such emotion, which in both cases produces the highest reported variance. This observation might be partially explained by sadness being the least frequent emotion, as demonstrated by Figure \ref{fig:db_em_dist}.

The described results strongly suggest that the model can successfully learn to simultaneously classify a dream report with respect to references to the different emotions of the HVDC coding system. However, the achieved performance level might be mediated, at least in part, by specific series of DreamBank. It is in fact possible that different emotions distribute in a particular and unique way in each series. If so, the model could learn series-specific distributions, and, after implicitly recognising a specific series in a given report, simply reproduce these distributions at test time. For example, if a series like \texttt{Ed} contained a large number of reports labelled both with \textit{sadness} and \textit{apprehension}, the model could implicitly learn to identify \texttt{Ed}'s reports from such series via recurrent cues to unrelated information (such as characters or places) and, at test time, use these cues to automatically annotate those reports with textit{sadness} and \textit{apprehension}.

\subsubsection{Ablation}\label{sec:ablation}
To understand whether the performance of the trained model is affected by this heuristical behaviour --- that is, learning series-specific emotion distributions --- we conduct a follow-up ablation experiment. Using the same architecture, hyper-parameters, and training setup, instead of randomly splitting five times the whole dataset into an 80-20\% train-test split, we here use one whole series of the dataset as the test set and the remaining series as the training set. With this approach, test series are never seen by the model during training, making it impossible for the model to rely on series-specific associations for solving the task. For this experiment, we focus solely on the General Emotions set, found to be the best-performing and more stable set. Moreover, we focus the analysis on the F1 scores as the performance metric of choice.

\begin{figure}[htb!]
    \centering
    \includegraphics[width=\columnwidth]{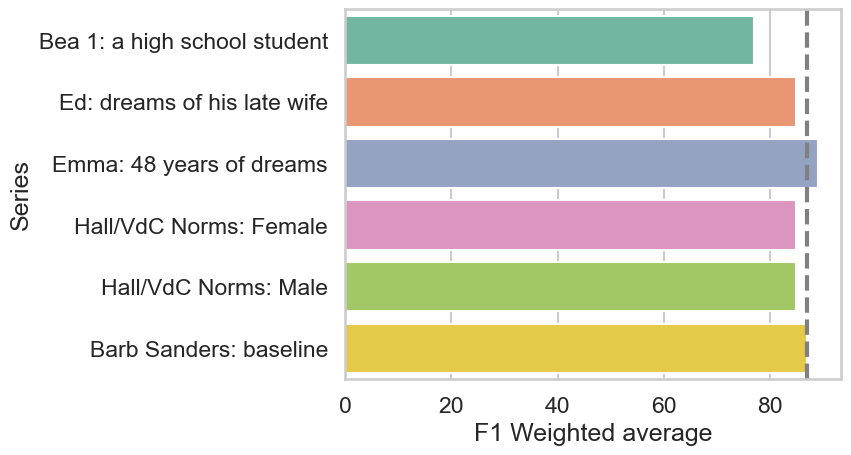}
    \caption{\label{fig:textclass_coll_general}Ablation experiment results. F1 Weighted average scores obtained by the model when each Dream Bank's Series is held out of training and used as a test set. The dashed vertical line reports the average F1 Weighted average obtained in the main experiment (see Table \ref{tab:multi_class_kfold_overall}). }
\end{figure}  

\begin{figure}[htb!]
    \centering
    \includegraphics[width=\linewidth]{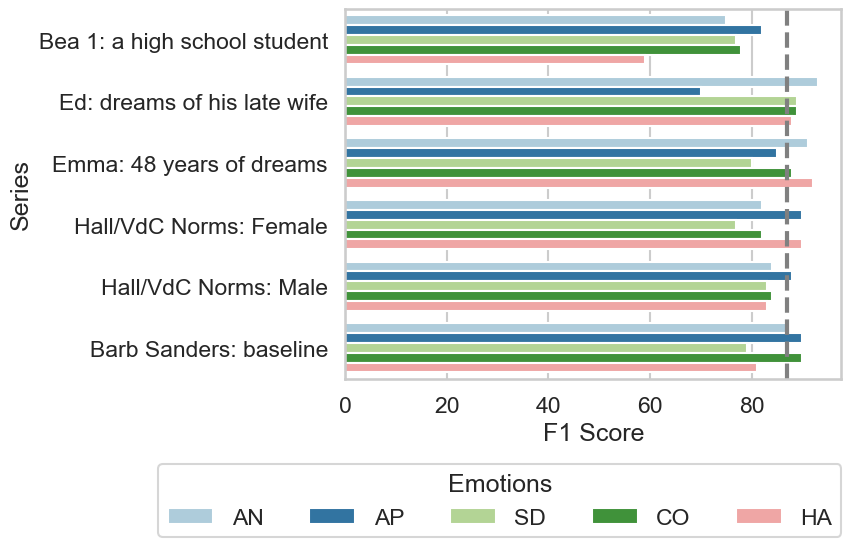}
    \caption{\label{fig:ablation_emotion_series}Ablation experiment, emotion by Series analysis. The diagram further breaks down the results of Figure \ref{fig:textclass_coll_general} by single emotion for each Series held out from training. Once again, the vertical dotted line refers to average scores in the main experiment (see Table \ref{tab:multi_class_kfold_overall}).}
\end{figure}

Figure \ref{fig:textclass_coll_general} summarises the results of the ablation experiment. The x axis shows the F1 weighted average scores obtained for each series (y axis) when such a series was held out from training and used as the test set. In order to facilitate comparison with the previous experiment’s results, the dashed grey line indicates the F1 weighted average obtained in the $K$-fold experiment (i.e., 87 ± 2, see Table \ref{tab:multi_class_kfold_overall}). The results indicate that when all the instances of a series are removed from the training data, the test performance of the model remains relatively high and stable. Moreover, as shown in Figure \ref{fig:ablation_emotion_series} this remains true for all of the HVDC scored emotions.

\begin{figure}[htb!]
    \centering
    \includegraphics[width=\linewidth]{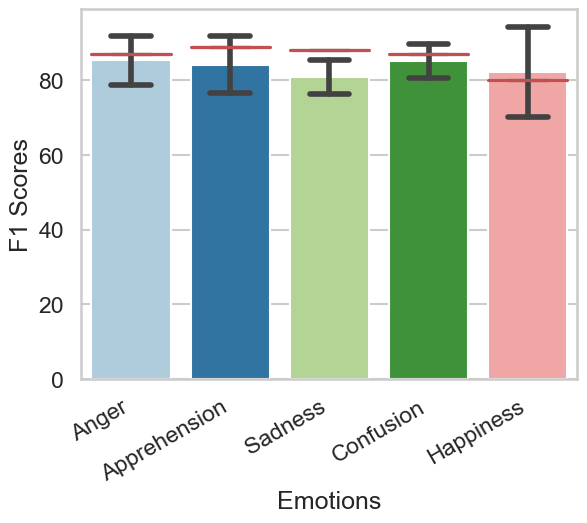}
    \caption{\label{fig:ablation_emotion}Ablation experiment, single emotion analysis. Overall results (in F1 scores) for every single emotion obtained in the different Series for the ablation experiment (see Figure \ref{fig:ablation_emotion_series}). Bars report standard deviation, while Red horizontal lines refer to average scores in the main experiment (see Table \ref{tab:multi_class_kfold_overall}).}
\end{figure}

The \texttt{Bea 1} series, appeared to represent the only notable exception to the above observations. Indeed, this series shows the greatest deviation from the original results, with an F1 weighted average of 77, compared to the previously obtained average of 87 (± 2). As shown in Figure \ref{fig:ablation_emotion_series}, which breaks down the results of the ablation experiment presented in Figure \ref{fig:textclass_coll_general} by single emotions, this was largely due to a problematic classification of \textit{happiness} (HA) in this particular series. However, with the exception of a slightly lower \textit{confusion}, emotions don't seem to significantly deviate from the $K$-fold experiment results, as summarised by Figure \ref{fig:ablation_emotion}.

The above results support two main conclusions. First, the proposed architecture, based on a pre-trained LLM, can learn efficient classification strategies for dream reports' emotional content, based on the HVDC coding system. Second, the learned model does not rely on simple heuristics based on series-dependent cues and distributions. 


The above results support two main conclusions. First, the proposed architecture, based on a pre-trained LLM, can learn efficient classification strategies for dream reports' emotional content (as defined based on the HVDC coding system). Second, the learned model does not rely on simple heuristics based on series-dependent cues and distributions. However, as currently formulated, the ablation results could have been influenced by yet another confound, based on the fact that different series have very different numbers of reports and emotion distributions (see Figure \ref{fig:db_em_s_dist}). In other words, the performance of each combination of series-emotion (e.g., \texttt{Bea 1}-\textit{happiness}) could be explained by the number of items provided at test time. To assess this possibility we performed a set of series-independent Spearman's correlations between the number of test items for each emotion and their respective results (i.e., the F1 scores). The results, summarised in Appendix \ref{app_ablation_correlation} found no connections between F1 scores and number of test items.

\subsubsection{Unlabelled-data analysis}
Although the set of available annotated reports is limited, DreamBank per se is a reasonably large resource, including many series of dream reports that have never been assessed using the HVDC coding system. Our proposed model could offer researchers an accessible and time-efficient method to evaluate references to emotions in non-scored series. To demonstrate this, we fine-tuned a model using the same architecture, paramenters, and procedure thus far adopted, with the only addition of an early-stop mechanism and no $K$-fold. As a test-set, we adopt a DreamBank series containing dream reports from a Vietnam war veteran with post-traumatic stress disorder (PTSD)\footnote{Please note that, as mentioned in Section \ref{sec:task_benchmark}, this series was not included in the annotated data provided by \texttt{Dreambank.net}, and was obtained via the scraped version generated by Matt Bierner. Code and data for the scraper are available at \url{https://github.com/mattbierner/DreamScrape}, or, via Hugging Face datasets at \url{https://huggingface.co/datasets/DReAMy-lib/DreamBank-dreams}.}, who had frequent negative dreams and nightmares. In other words, while we do not have an \textit{actual} emotion distribution for such a series, we can \textit{assume} an expected one, with a strong predominance of negatively-connotated emotions. Code and the obtained model are made openly available here\footnote{\url{https://github.com/lorenzoscottb/Dream\_Reports\_Annotation/Experiments/Supervised\_Learning}}.

\begin{figure}[htb!]
    \centering
    \includegraphics[width=\linewidth]{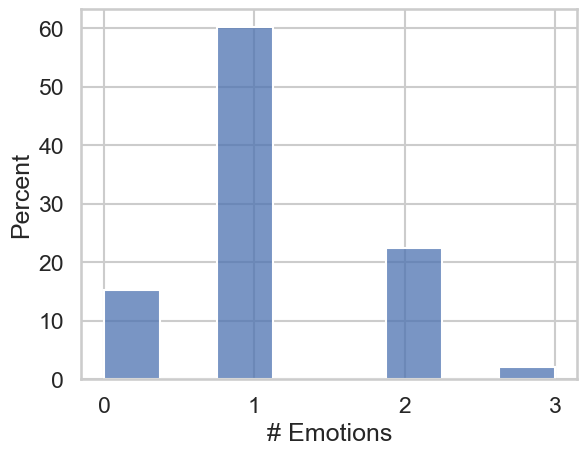}
    \caption{\label{fig:unlabel_vet_data} Number of predicted emotions per report. Distribution of the number of emotions-per-report predicted by the model for the \texttt{Veteran} series.}
\end{figure}  

\begin{figure}[htb!]
    \centering
    \includegraphics[width=\linewidth]{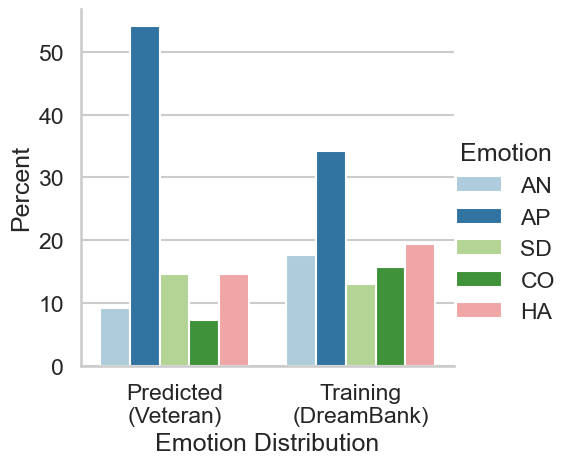} \caption{\label{fig:db_em_dveteran_db_emot_ditist} \texttt{Veteran} and DreamBank emotion distributions. Comparison of the emotion-distribution predicted by the model for the (unlabelled) \texttt{Veteran} series, and the general emotion distribution in the DreamBank dataset, used to train the model.)}
\end{figure}

Out of the 98 dreams contained in the \texttt{Veteran} series, the model found at least one emotion in approximately 84\% of them. As summarised by Figure \ref{fig:unlabel_vet_data}, most of these reports were associated by the model with a single emotion, and approximately 20\% of them were labelled with two emotions. As expected, the vast majority of these reports contain negatively connotated emotions, as seen in Figure \ref{fig:db_em_dveteran_db_emot_ditist}. \textit{Apprehension} is by far the most observed negative emotion, appearing in more than half of these reports. Moreover, Figure \ref{fig:db_em_dveteran_db_emot_ditist} strongly suggests that the emotion distribution proposed by the model for the \texttt{Veteran} series is not simply a transposition of the one observed by the model during training. This further suggests that the model has successfully learned reliable and generalised classification strategies, rather than simply reproducing observed distribution from the training data.

Of note, the model also identified a minority of reports – circa 19\% – that are believed to contain (also) happiness. A manual inspection of this small set did identify some errors, but also found multiple instances where the model’s annotation (i.e., including happiness as an emotion expressed within the report) seemed justified. For example, in one of these reports, after describing a very violent war scenario, the Veteran adds that he felt \textit{``a feeling of complete freedom. In very high spirits Jim L. and I go to a supermarket and buy food. I am aware that I don't wear my steel helmet. ''}. In another case, a dream that would be easily classified as a nightmare (two dolls have come to life) is narrated in a  normal and friendly manner, as clear from the passages like \textit{``I speak to the male doll and the female doll and feel happy. I have made two friends.''}.

In other cases, the negative context of the dream is notably less dominant, and the report simply describes a series of social encounters and interactions. Lastly, we found more straightforward classifications, clearly triggered both by strong cue words and context, like happy/happiness. In one report, for instance, the Veteran is in a rehab clinic, surrounded by other veterans, and painting from children presents as \textit{``[...] colourful, lively, happy. There is no sense of war''}; in another, he describes a romantic encounter with a woman – \textit{``We are happy and young. She puts her arm around my shoulder. " I like you, " she says. " I really like you."''}.




\section{Discussion}
In the field of dream research, the assessment of references to emotions in dream reports is typically based on time-consuming, operator-dependent procedures. Throughout the years, only a few studies employ automated approaches based on NLP techniques, including dictionary-based and distributional semantics methods. However, these approaches have very limited access to the syntax and semantics of a report's content, and may thus fail to correctly and fully capture emotions described in dream reports. In this work, we tested whether transformer-based large language models (LLM) could be used to overcome such a limitation and potentially replace human-based scoring in dream research. Experiments were focused on two main approaches. First, we used an off-the-shelf LLM tuned to perform a binary (\texttt{POSITIVE} vs. \texttt{NEGATIVE}) sentiment analysis. The obtained results showed that the selected off-the-shelf model failed to capture the variability of dream reports' general sentiment. While the model's predictions were extremely confident, and, hence, clustered around the extremes of the decision boundary, observed reports' sentiments were more smoothly and normally distributed. Moreover, the analyses showed that the performance of the model greatly varied across the different subsets (or series) of the DreamBank dataset. The second approach was based on full supervision. Specifically, we trained a model end-to-end using pre-annotated data to predict if and which emotions were present or absent in dream reports. The obtained results showed that the model was able to learn reliable and stable classification rules. An ablation experiment further confirmed that the ability of the model to solve the task was only marginally affected by differences between distinct data series, as observed instead for the off-the-shelf model.

The described observations suggest that what is more likely to impact the model performance is the vocabulary used to describe specific emotions across different series. Indeed, variability in the used vocabulary may be explained by the fact that the series included in the present work were collected from different individuals or groups of individuals, with relevant differences in demographic, psychological, and behavioural characteristics. If correct, this observation would be another reason to support the use of NLP tools that are able to reason on the full content of a report, have access to a large vocabulary, already have significant information about a large set of lexemes, and can be easily adaptable to new words. Pre-trained transformer-based large language models satisfy all these requirements. Given the current state of NLP tools and resources, our proposed architecture can be adapted to be used with an LLMs pre-trained on different languages or tasks, and even use a multi-lingual one. Trained models like ours are fully open-source and can be easily uploaded and downloaded, making the results and the framework extremely replicable and widely standardised. 

It is important to acknowledge that this study has two main limitations. First, while DreamBank does contain reports in multiple languages, the HVDC annotations were available only for reports in English. Thus generalizability of our observations to other languages remains unclear. Second, while the analyzed dataset was relatively large with respect to published studies in the field of dream research, it was instead relatively small for a machine-learning investigation, and especially for the use of supervised methods.

\section{Conclusion}
In this work, we tested the feasibility of using transformer-based large language models (LLM) to annotate dream reports with respect to references to experienced emotions. Our results show that, while an off-the-shelf LLM tuned on sentiment analysis struggles to replicate the general sentiment of a report, and is heavily impacted by internal differences in the dataset, our supervised solution based on multi-label classification yields a strong performance, which was found to be robustness with respect to the internal differences of the dataset that afflicted the unsupervised solution. Such approaches have the potential to significantly accelerate research investigating the origin, meaning, and functions of dreams, and might present a valuable and efficient support or alternative to human-based procedures involving the analysis of large datasets, ensuring at the same time reproducibility of the obtained results through the sharing of adopted models.




\section*{Acknowledgements} 
This research was supported by the EU Horizon 2020 project HumanE-AI (grant no. 952026), and a BIAL Foundation Grant (grant no. 091/2020). We thank Ian Morgan Leo Pennock and Giacomo Handjaras for their comments on an early draft of the work.




\appendix

\section{DreamBank's Distributions}\label{app_DB_dist}
\begin{figure}[htb!]
    \centering
    \includegraphics[width=\linewidth]{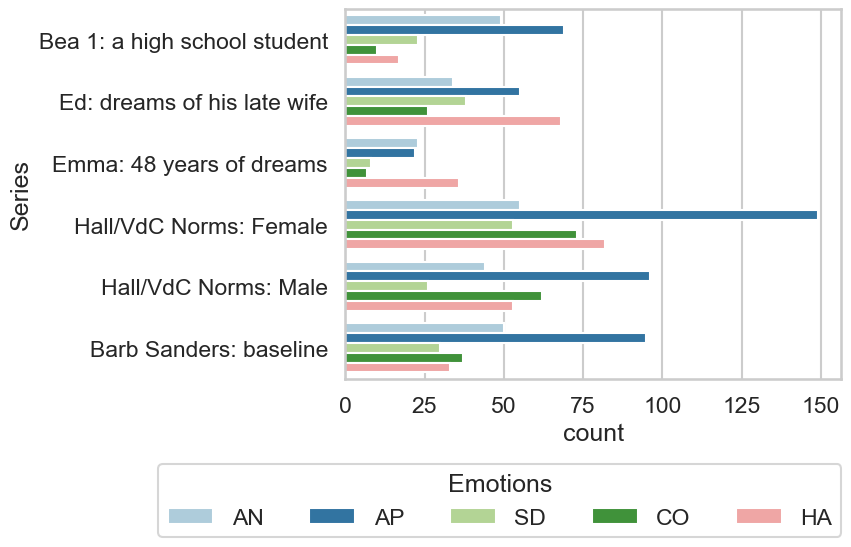} \caption{\label{fig:db_em_s_dist} General emotion distribution across Dream Bank's Series.}
\end{figure}

\begin{figure}[htb!]
    \centering
    \includegraphics[width=\linewidth]{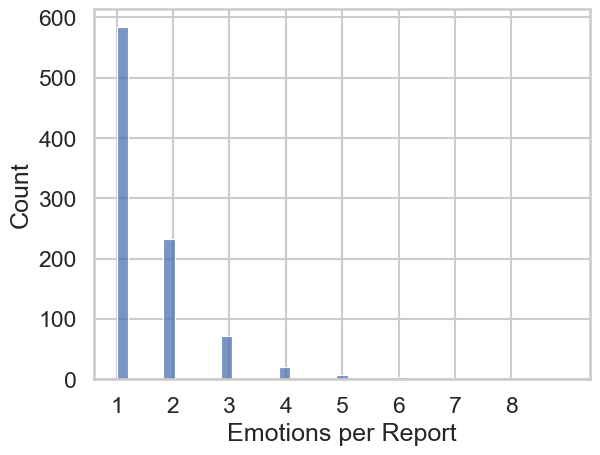} \caption{\label{fig:db_em_report_dist} Number of emotion per report. Visualisation of how reports distribute with respect to the number of (General) emotions they have been labelled with.}
\end{figure}

\begin{figure}[htb!]
    \centering
    \includegraphics[width=\linewidth]{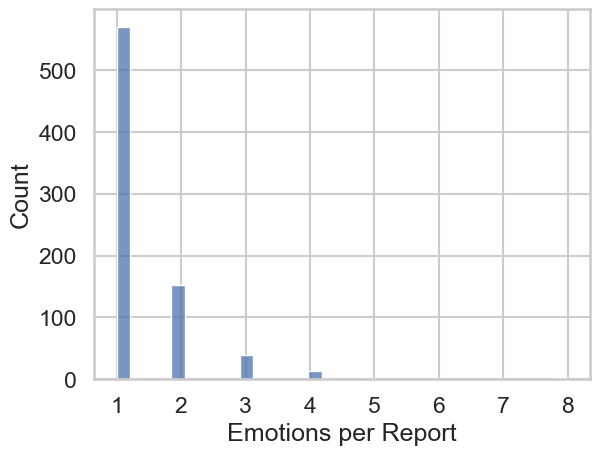}\caption{\label{fig:db_dreamer_em_report_dist} Number of Dreamer-only emotion per report. Visualisation of how reports distribute with respect to the number of (Dreamer-only) emotions they have been labelled with.}
\end{figure}

The current section proposes more details and analyses of DreamBank's statistics. Figure \ref{fig:db_em_s_dist} presents the distribution of the HVD emotions in DreamBank, divided between the different series of DreamBank.  Figure \ref{fig:db_em_report_dist} presents how single DreamBank reports distribute with respect to the number of (General) emotions per report. As shown, the majority (circa 65\%) of the 922 reports containing at least one emotion in fact contain only one emotion. Approximately 25\% contains two emotions, while the rest can reach up to 9 emotions per report. As hinted by Figure \ref{fig:db_dreamer_em_report_dist}, the percentage of reports with only one emotion reaches almost 75\%, bringing the number of reports with more than two emotions to circa 5\% of the total.

\section{Annotator vs. Model Scores Analysis}\label{app_hists}
\begin{figure}[htb!]
    \centering
    \includegraphics[width=\columnwidth]{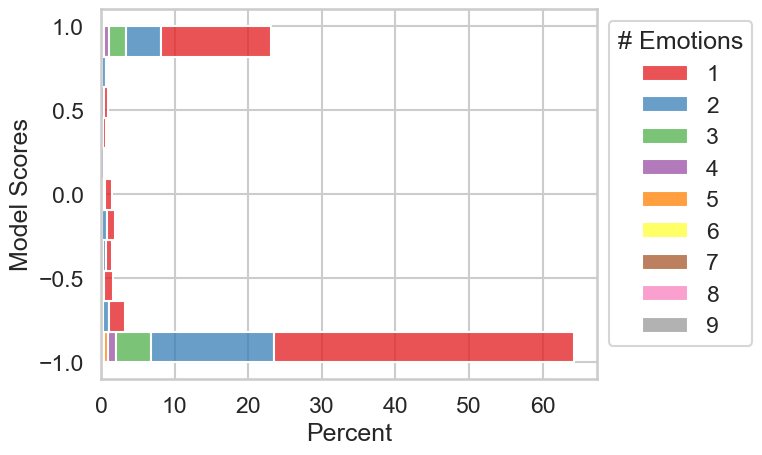}
    \caption{\label{fig:model_scores_hist}\texttt{Model scores} distribution. In detail visualisation of the \texttt{Model scores} distribution, divided by the number of emotions per report presented in Figure \ref{fig:general_sentiment_main} from Section \ref{sec:usa_RS}.}
\end{figure}

\begin{figure}[htb!]
    \centering
    \includegraphics[width=\columnwidth]{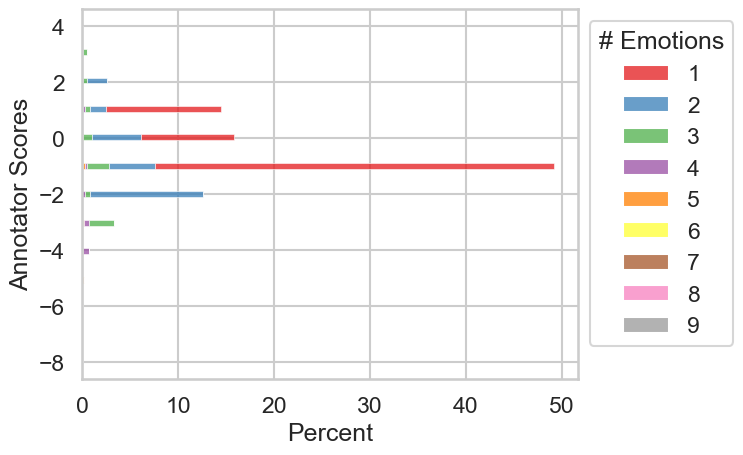}
    \caption{\label{fig:annotator_scores_hist}\texttt{Annotator scores} distribution. In detail visualisation of the \texttt{Annotator scores} distribution, divided by the number of emotions per report presented in Figure \ref{fig:general_sentiment_main} from Section \ref{sec:usa_RS}.}
\end{figure}

The section presents a more detailed analysis of the Model and Annotator score distributions, with respect to the number (\#) of emotions. As shown in Figure \ref{fig:model_scores_hist}, the two peaks of the \texttt{Model Scores}distributions mainly contain reports classified by annotators as presenting a single emotion. However, the proportion of reports containing two emotions is notably higher in those reports classified by the model as being strongly \texttt{NEGATIVE}. Interestingly, with respect to the \texttt{Annotator Scores}, the proportion of reports with two emotions is concentrated in those reports with \texttt{Annotator Scores} of –2 (see Figure \ref{fig:annotator_scores_hist})

\section{Supervised learning hyper-parameters}\label{app_hp}
Description of the hyper-parameters used to train the architectures and models described throughout Section \ref{sec:classification}. For the relevant code see here\footnote{\url{https://github.com/lorenzoscottb/Dream_Reports_Annotation/tree/main/Experiments/Supervised_Learning}}.

\begin{table}[htb!]
\centering
\resizebox{.7\columnwidth}{!}{%
\begin{tabular}{lc}
\hlineB{3}
\textbf{Parameter} &    \textbf{Value} \\
\hlineB{3}
BERT-input max-len & 512 \\
epochs & 10 \\
learning rate & 0.00001 \\
batch size & 8 \\
input truncation & True \\ 
truncation-to & max-length \\ 
\hlineB{3}
\end{tabular}
}
\caption{\label{tab:supervised_hp} Hyper-parameters used for training the architectures in Section \ref{sec:classification}.}
\end{table}

\section{Support-score correlation analysis}\label{app_ablation_correlation}
\begin{figure}[htb!]
    \centering
    \includegraphics[width=\linewidth]{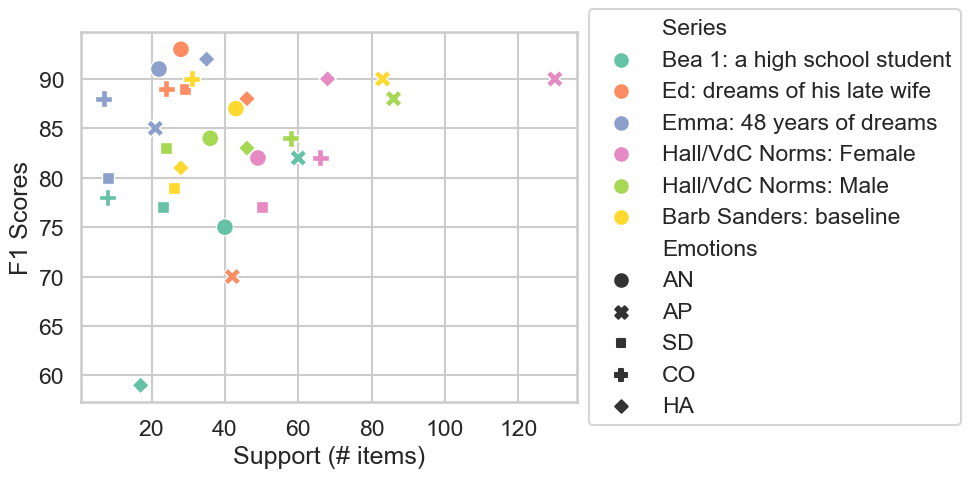}
\caption{\label{fig:ablation_emotion_series_support}Ablation's experiment, score vs support correlation analysis. Visualisation of the correlation analysis, presented in Table \ref{tab:support_correlation}, between the number of test items (x axis) and F1 scores (y axis), for each combination of Series and emotion.}
\end{figure}

Table \ref{tab:support_correlation} and Figure \ref{fig:ablation_emotion_series_support} summarise the results of the correlation analysis from Section \ref{sec:ablation} summarised in T. Overall, this analysis indicated no clear connection between the number of test instances containing a specific emotion and the models' final performance on the ablation experiment.


\begin{table}[htb!]
\centering
\resizebox{\linewidth}{!}{%
\begin{tabular}{lcc}
\hlineB{3}
\textbf{Series} &  \textbf{Spearman's $\rho$} &    \textbf{\textit{p}}  \\
\hline
Bea 1: a high school student &        0.3000 &  0.6238  \\
Ed: dreams of his late wife &       -0.7182 &  0.1718 \\
Emma: 48 years of dreams &        0.7000 &  0.1881  \\
Hall/VdC Norms: Female &        0.7906 &  0.1114 \\
Hall/VdC Norms: Male &        0.7379 &  0.1546 \\
Barb Sanders: baseline &        0.8208 &  0.0886\\
\hlineB{3}
\end{tabular}
}
\caption{\label{tab:support_correlation} Correlation analysis between F1 score and support (\# items) per single emotion in the ablation experiment. Each row of the table presents the results of the correlations between the number of instances containing a given emotion, and the obtained F1 scores (see Figures \ref{fig:ablation_emotion_series} and \ref{fig:ablation_emotion_series_support} for further visual breakdowns). Columns describe the single Series under investigation, the $\rho$ coefficient and the $p$ value of each correlation.}
\end{table}


\bibliography{tacl2021, anthology}
\bibliographystyle{acl_natbib}

\end{document}